%% file: aaai24.tex
\documentclass[letterpaper]{article} 
\usepackage{aaai24}  
\usepackage{times}  
\usepackage{helvet}  
\usepackage{courier}  
\usepackage[hyphens]{url}  
\usepackage{graphicx} 
\usepackage{natbib}  
\usepackage{caption} 
\usepackage{algorithm}
\usepackage{algorithmic}

\usepackage{newfloat}
\usepackage{listings}
\DeclareCaptionStyle{ruled}{labelfont=normalfont,labelsep=colon,strut=off} 
\floatstyle{ruled}
\newfloat{listing}{tb}{lst}{}
\floatname{listing}{Listing}
\usepackage{graphicx}
\usepackage{amsmath}
\usepackage{amssymb}
\usepackage{booktabs}

\usepackage{multirow}
\usepackage{array}
\usepackage{color}
\usepackage{xcolor}
\usepackage{subcaption}
\usepackage{pifont}
\usepackage{makecell, booktabs}

\newcommand{\methodname}{\textsc{VLN-Video}}

\title{\methodname{}: Utilizing Driving Videos for Outdoor Vision-and-Language Navigation}

\author {
    Jialu Li\textsuperscript{\rm 1}\footnote{Work done during an internship at Amazon.},
    Aishwarya Padmakumar\textsuperscript{\rm 2},
    Gaurav Sukhatme\textsuperscript{\rm 2},
    Mohit Bansal\textsuperscript{\rm 1}
}
\affiliations {
    \textsuperscript{\rm 1}University of North Carolina, Chapel Hill\\
    \textsuperscript{\rm 2}Amazon Alexa AI\\
    \{jialuli, mbansal\}@cs.unc.edu,
    \{padmakua, sukhatme\}@amazon.com
}

\usepackage{bibentry}

\begin{document}

\maketitle

\begin{abstract}
Outdoor Vision-and-Language Navigation (VLN) requires an agent to navigate through realistic 3D outdoor environments based on natural language instructions. The performance of existing VLN methods is limited by insufficient diversity in navigation environments and limited training data.
To address these issues, we propose \methodname{}, which utilizes the diverse outdoor environments present in driving videos in multiple cities in the U.S. augmented with automatically generated navigation instructions and actions to improve outdoor VLN performance.
\methodname{} combines the best of intuitive classical approaches and modern deep learning techniques, using template infilling to generate grounded navigation instructions, combined with an image rotation similarity based navigation action predictor to obtain VLN style data from driving videos for pretraining deep learning VLN models. 
We pre-train the model on the Touchdown dataset and our video-augmented dataset created from driving videos with three proxy tasks: Masked Language Modeling, Instruction and Trajectory Matching, and Next Action Prediction, so as to learn temporally-aware and visually-aligned instruction representations. 
The learned instruction representation is adapted to the state-of-the-art navigator when fine-tuning on the Touchdown dataset. Empirical results demonstrate that \methodname{} significantly outperforms previous state-of-the-art models by 2.1\% in task completion rate, achieving a new state-of-the-art on the Touchdown dataset. 
\end{abstract}

\input{introduction}
\input{related_work}
\input{generate_data}
\input{model_training}
\input{experiments}
\input{conclusion}

\bibliography{aaai24}

\appendix
\input{Supplementary}

\end{document}

%% file: introduction.tex
\section{Introduction}
\label{sec:intro}

\begin{figure*}[t]
\begin{center}
\includegraphics[width=1.0\linewidth]{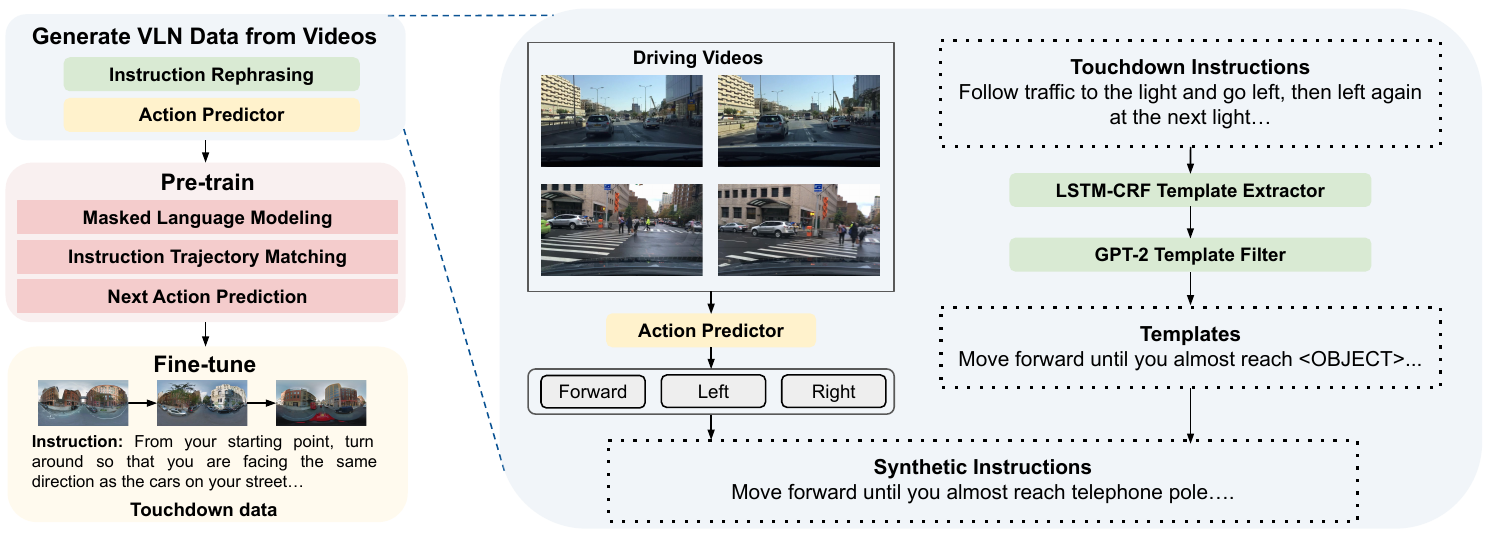}
\end{center}
  \caption{Overview of our proposed method \methodname{}: We annotate driving videos with synthetic navigation instructions by extracting instruction templates from the Touchdown dataset and filling them with actions predicted using our image rotation similarity based navigation action predictor and objects detected using a pre-trained object detector. We pre-train VLN models on both the processed video data and Touchdown data to learn better domain knowledge with three proxy tasks. Lastly, we transfer the learned language representation to VLN via  fine-tuning. 
}
\label{figure1}
\end{figure*}

Vision-and-Language Navigation (VLN) requires an agent to navigate through 3D environments based on natural language instructions~\cite{brahmbhatt2017deepnav, mirowski2018learning, anderson2018room2room, misra2018mapping, blukis2020learning, thomason:corl19, nguyen2019hanna,  Chen19:touchdown, ALFRED20, qi2019rerere, hermann2019learning, berg2020grounding, ku2020room, mehta2020retouchdown}. A critical bottleneck in improving models for VLN tasks is the limited availability of training data. Training data for VLN are usually collected with human annotation, where multiple annotators write instructions for a sampled trajectory in the environment. Additional annotators read these instructions and navigate accordingly to evaluate their quality and clarity. This annotation process is expensive and time consuming, which makes it hard to collect a large dataset for VLN. 

Many methods have been proposed to solve the data scarcity problem in VLN. For indoor VLN, several works leverage pre-trained multi-modal representations from large datasets such as BookCorpus~\cite{Zhu_2015_ICCV} for natural language and Conceptual Captions~\cite{sharma2018conceptual} for images, and further enhance the agent with specific domain knowledge by pre-training on an in-domain navigation dataset \cite{hao2020towards, majumdar2020improving, chen2021history, qiao2022hop}, sometimes enhanced with additional out of domain images~\cite{guhur2021airbert, li2022envedit, liu2021vision, chen2022learning}.  
Other works perform multimodal augmentation by training a speaker model to generate synthetic navigation instructions for unlabeled trajectories~\cite{hao2020towards, he2021landmark, dou2022foam, fried2018speaker, tan2019learning}.
Such augmentation methods can result in promising performance gains but prior results have focused primarily on indoor environments, in contrast to our focus on outdoor VLN.
Additionally, most of these approaches do not introduce new training environments to improve generalization to unseen environments.  
Some add out of domain images~\cite{guhur2021airbert} but they lack the temporal information that could help an agent learn causality between actions and consequent observations.
To address these problems, we propose to utilize large video datasets that have diverse new environments and both spatial and temporal information for pre-training to further enhance the ability of trained agents to reason.
Additionally, if we can effectively utilize video for pretraining VLN agents, there exist a number of other video datasets ~\cite{chen2018bdd100k, grauman2022ego4d, pirsiavash2012detecting, caba2015activitynet, abu2016youtube} that could be potentially useful for indoor or outdoor VLN tasks.

In this work, we propose \methodname{}, a new data augmentation method that enhances outdoor VLN by learning from driving videos (Figure~\ref{figure1}). 
We process driving videos from the BDD100k dataset~\cite{chen2018bdd100k} to obtain pre-training data for the Touchdown dataset~\cite{Chen19:touchdown}.
To utilize driving videos for pre-training, we need to generate language navigation instructions for each video and predict navigation actions - move forward, turn left or turn right - between each pair of consecutive frames. Learning the instruction generator is challenging for three reasons. First, the entities referenced in outdoor navigation follows a long-tail distribution, where an object such as ``intersection" appears in 29\% of the sentences in the instructions. This when combined with the small size of the training set and long average instruction length of 89.6 words causes a speaker model to overfit to frequently occurring objects and generate repetitive instructions. Further, the entities mentioned in the instruction only take a small portion of the observation and are unlikely to be centered in the image, which make it difficult to learn the alignment between observations and instructions. 
Finally, the large domain gap between Touchdown images in Manhattan and BDD100K videos in a larger number of cities makes transferring a speaker model challenging.

Due to the above challenges of training speaker models in our setting, we propose a template based method for generating navigation instructions - masking out noun phrases and navigation directions from the original instructions, and filling them with select objects detected in the observation by a pre-trained object detector and navigation actions predicted using an intuitive image rotation similarity based predictor.

With the augumented data from videos, we pre-train the agent with three proxy tasks: Masked Language Modeling, Instruction and Trajectory Matching, and Next Action Prediction to learn temporally-aware and visually-aligned instruction representations. We adapt the learned instruction representations to the state-of-the-art navigation agent and fine-tune it on the Touchdown dataset.

We show that \methodname{} significantly improves over non-pre-training baselines by 2.1\% in task completion rate (TC) on the Touchdown~\cite{Chen19:touchdown} test set, achieving the new state-of-the-art for Touchdown.
We improve over pre-training with only in-domain data by 2.9\% TC on the validation set and demonstrate that pre-training on the synthetic data generated with our template infilling method on StreetLearn dataset achieves better performance than the style-transferred Google Maps instructions~\cite{zhu-etal-2021-multimodal}. We also show that our template infilling method and rotation similarity based action predictor work better than learning based methods, improving the performance by 7.4\% and 3.2\% TC respectively. Finally, we qualitatively show that our generated instructions align with the trajectory better. 

%% file: related_work.tex
\section{Related Work}
\noindent\textbf{Vision-and-Language Navigation.} 
Vision-and-Language Navigation is a task that requires an agent to navigate through a 3D environment based on natural language instructions and egocentric visual observations. Many datasets have been proposed for this task~\cite{hermann2017grounded, mirowski2018learning, anderson2018room2room, misra2018mapping, thomason:corl19, nguyen2019hanna,  Chen19:touchdown, ALFRED20, qi2019rerere, hermann2019learning, ku2020room, mehta2020retouchdown}. While substantial progress has been made for indoor VLN~\cite{chen2021history, chen2022learning, qiao2022hop, kim2021ndh, wang2019reinforced, hao2020towards, zhou2021rethinking, zhang2022explicit, lin2022adapt, lin2022multimodal, georgakis2022cross, chen2022weakly, fu2020counterfactual}, 
outdoor VLN is still an under-explored area~\cite{zhu-etal-2021-multimodal, schumann2022analyzing, xiang2020learning, armitage2022priority}. 
The dataset most commonly used to study language guided outdoor navigation is the Touchdown dataset~\cite{Chen19:touchdown} which involves navigation instructions set in Manhattan, with StreetView panoramas as observations.
Prior work on this dataset includes an LSTM navigation agent with cross modal attention to ground instructions in navigation history~\cite{schumann2022analyzing}, using trajectory traces as external knowledge to aid agents' navigation~\cite{armitage2022priority}, and using a transformer based architecture combined with a data augmentation method that transfers the style of the instructions in the StreetLearn dataset~\cite{mirowski2019streetlearn} subsequently used for directly pre-training the agent with the downstream navigation task~\cite{zhu-etal-2021-multimodal}.
We explore a new data source -  driving videos - for data augmentation to enrich the navigation environments, which requires novel techniques for instruction generation and navigation action prediction.

\noindent\textbf{Data Augmentation in Vision-and-Language Navigation.}
Data scarcity is a core problem in VLN. There exists two ways of data augmentation: instruction-level data augmentation or environment-level data augmentation. The former generates synthetic instructions for unannotated paths in the existing navigation environments, while the latter tries to create new navigation environments. 
Many works attempt to learn deep learning based speaker models to generate synthetic instructions for unannotated paths in the Matterport 3D~\cite{chang2017matterport3d} environments to improve performance on indoor VLN datasets set in these environments~\cite{tan2019learning, hao2020towards, fried2018speaker, dou2022foam, he2021landmark, zhao2021evaluation}.
Recent research aims to augment indoor navigation environments by mixing~\cite{liu2021vision} or editing~\cite{li2022envedit} existing environments and utilizing external room images~\cite{guhur2021airbert, chen2022learning}. 
For outdoor VLN, prior work transfers the style of the instructions in the StreetLearn dataset~\cite{mirowski2019streetlearn} to match the instruction style on Touchdown dataset~\cite{zhu-etal-2021-multimodal}. This also provides an augmented navigation environment as StreetLearn covers a larger geographical area compared to Touchdown.
However, their approach is dependent on the existence of ground truth navigation instructions in the dataset used for augmentation, which is not the case for other data sources such as outdoor/city images or driving videos from the web. \methodname{} proposes an instruction generation method that does not rely on ground truth navigation instructions, and is the first to explore the utilization of videos as a source for augmented data in outdoor VLN.

\noindent\textbf{Utilizing Videos in Pre-training.}
Videos have been utilized during pre-training for downstream tasks such as text-to-video retrieval~\cite{xu2016msr, krishna2017dense, rohrbach2015dataset}, video question answering~\cite{xu2017video, jang2017tgif, lei2018tvqa}, and video classification~\cite{ goyal2017something, kay2017kinetics}. Recently, videos have been used in pre-training for non-video tasks such as playing a video game - Minecraft~\cite{baker2022video}, and pre-training the policy representation for a driving agent~\cite{zhang2learning}. 
However, these works focus on uni-modal tasks. Our paper is the first to explore utilizing videos during pre-training for multi-modal instruction-guided navigation tasks.

%% file: generate_data.tex
\section{Generating VLN Data from Driving Videos}

\begin{figure*}[t]
\begin{center}
\includegraphics[width=1.0\linewidth]{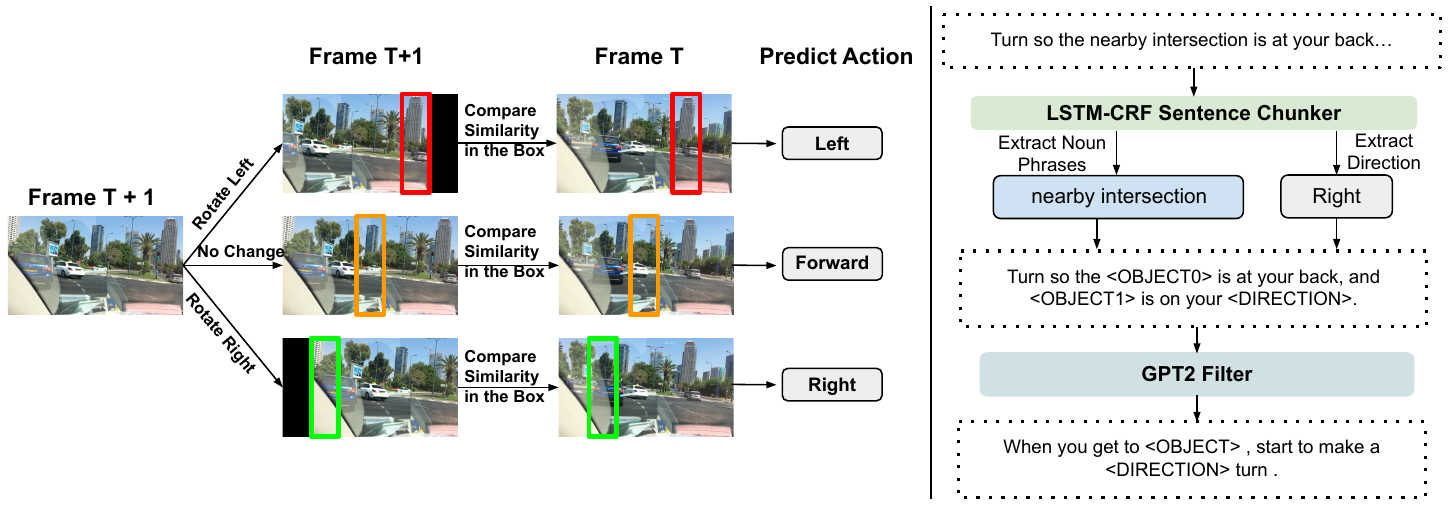}
\end{center}
  \caption{
  Overview of our proposed method to detect the turn between consecutive frames in driving videos, and generate synthetic instructions with template infilling method.
}
\label{figure2}
\end{figure*}

In this section, first we explain the challenges in generating synthetic instructions for videos data in Sec.~\ref{difficulty}. Next, we describe our approach that automatically generates large-scale VLN data from caption-less driving videos. Specifically, we use driving videos from the training set in BDD100k~\cite{chen2018bdd100k}. Our template infilling based generation approach contains three parts: instruction template extractor, image rotation similarity based navigation action predictor, and object detector. We first extract instruction templates from the training data in the Touchdown dataset, where we mask out direction words and entity references in the original instructions (Sec.~\ref{template_generation}). 
We then predict navigation actions between consecutive frames (Sec.~\ref{action_prediction}), detect objects in the current observation (Sec.~\ref{obj_detector}), and use these to fill templates to obtain navigation instructions (Sec.~\ref{ins_generation}).

\subsection{Challenges of Instruction Generation from Videos} \label{difficulty}

Although many works have explored generating synthetic instructions from navigation trajectories for indoor VLN using deep learning models, such models fail the outdoor Touchdown dataset for three reasons: 

\noindent\textbf{Difficulty in Aligning Objects in Observations with Entities in Instructions.} 
The objects referenced by entities in outdoor VLN instructions usually take up a small area in the full panoramic outdoor image observation. This makes such objects less likely to be detected by deep learning object detectors and more difficult for a deep learning model to learn the cross-modality grounding between the instruction and the observation, which is crucial for learning a good instruction generator.  
VLN models trained on these datasets are also found to rely less on detected objects - it is shown that even when 100\% of the objects are masked out in the instruction, an LSTM VLN model's performance only drops by less than 3\% in task completion~\cite{schumann2022analyzing}.
Further, although the navigation instructions in Touchdown~\cite{Chen19:touchdown} have rich references to entities in each instruction (10.7 entities on average~\cite{Chen19:touchdown}), these follow a long-tail distribution.
For example, ``intersection" appears 2.1 times on average in each training instruction, and 29.0\% of the sentences in the dataset contain the entity ``intersection".
This imbalance hinders the ability of a speaker model to learn the alignment between observations and instructions based on object information, and causes speaker models to overfit to objects that appear frequently.   

\noindent\textbf{Large Environment Gap between Navigation Environment and Driving Videos.} There usually exists little to no gap between the training and inference environments for a speaker model for indoor VLN. 
Most works use the speaker model to generate synthetic instructions on un-annotated paths in the same environments~\cite{tan2019learning, hao2020towards, dou2022foam, fried2018speaker, he2021landmark}. 
Given the large gap between the panorama observations in Touchdown environments and egocentric views in the driving videos, the performance of directly adapting a speaker model trained on Touchdown environments to driving videos is very low. Although the large domain gap poses challenges in generating synthetic instructions for videos, it also has the potential to enrich the navigation environments with more diverse visual observations from driving videos. 

\noindent\textbf{Long Instructions and Trajectories in Outdoor Navigation.} 
The instructions in Touchdown are almost three times longer than the instructions in the indoor VLN dataset Room-to-Room~\cite{anderson2018room2room}, with an average length of 89.6 words. Further, the trajectories in Touchdown are nearly six times longer than the trajectories in Room-to-Room (35.2 vs. 6 steps on average). However, Touchdown only contains 6,526 training examples in contrast to 14,052 for Room-to-Room. 
The long instruction and trajectory length and limited available training data makes it hard to learn a speaker model that generates faithful navigation instructions.

We show in analysis that a speaker trained on Touchdown fails to generate high quality instructions for BDD100K videos, while our template infilling approach can generate instructions with rich entity mentions and little repetition.  

\subsection{Instruction Template Extraction} \label{template_generation}
We extract instruction templates from the training data of the Touchdown dataset. The template extraction method consists of two steps. First, we detect noun phrases and direction words in the sentence using an LSTM-CRF model with a pre-trained sequence labeling embedding~\cite{akbik2018coling} to chunk the instructions. 
The sequence labeling embedding is a contextualized character-level word embedding, which is learned by optimizing the next character prediction task. We mask noun phrases with $<$OBJECT$>$ tokens and filter out templates that have multiple $<$OBJECT$>$ tokens to reduce the probability of nonsensical sentences resulting from strange combinations of objects (for example we would want to avoid generating a sentence such as: $<$Street sign$>$ should be $<$intersection$>$ to your right). 
We then categorize templates into ``turn'', ``forward'' and ``stop'' templates by searching for the keywords ``right'', ``left'', ``forward'' and ``stop'' since humans tend to write instructions differently for each of these types.
We filter out templates that contain multiple direction words so that each template matches with one specific action. We then use GPT-2~\cite{radford2019language} to filter out templates with low generation probability. Specifically, for every template, we fill in the most frequently appearing objects (signboard, traffic light, awning, telephone pole) and the direction words (``left", ``right", ``forward", ``stop"), calculate its generation loss with GPT-2, and filter out templates with high generation loss. We filter out half of the templates with GPT-2. In total, we extract 3,004 templates for turn actions, 269 templates for forward actions, and 92 templates for stop actions.

\subsection{Image Rotation Similarity Based Navigation Action Prediction from Driving Videos} \label{action_prediction}
Touchdown~\cite{Chen19:touchdown} contains a graph of the navigation environments, where each node contains a panorama observation and its heading, and edges indicate connectivity between panoramas. Navigation actions can be derived from heading changes between two consecutive panoramas. However, video datasets do not have explicit heading annotation for each frame to derive navigation actions between consecutive frames. Thus, we need to generate actions for driving videos as pseudo labels for pre-training tasks such as next action prediction. We additionally utilize the predicted actions to generate more accurate instructions. 

One standard way to predict an action on the driving video dataset is to learn an action predictor on the Touchdown dataset. However, the action distribution in Touchdown is imbalanced with 91\% of the actions as ``forward'' actions. We experiment with three models for navigation action prediction: a multi-layer perceptron which takes in observations at two consecutive steps, an LSTM based decoder and a transformer based decoder that takes in the full navigation history. We also use under-sampling and weighted cross-entropy loss to mitigate the data imbalance. Though the best model achieves 82\% macro average accuracy on Touchdown validation data, when adapted to driving videos, they predict the ``forward'' action for almost all pairs of frames, likely due to the domain gap between Touchdown environments and driving videos in BDD100k. 

Thus, we propose a more intuitive and effective \emph{image rotation similarity} based method to predict navigation actions. We hypothesize that two consecutive frames in a video contain mostly the same information, which makes it hard for deep learning models to learn the small differences between consecutive frames to predict the action. 
Thus, we instead rotate the frame to mimic the car turning action, and compare the similarity between the rotated image and the target frame to determine the turning direction. As shown in Figure~\ref{figure2}, given two consecutive frames sampled from a video, we rotate the frame at time $t+1$ in both left and right directions. We hypothesize that if the frame rotated left has a higher similarity with frame $t$ than the frame rotated right or remain unchanged, it indicates that the car is turning left. Specifically, given a frame with size $H\times W$, and a window with size $D$, we calculate the similarity inside the window between rotated frame and frame $t$. The window will pass through the image from left to right, excluding the rotated part with width $R$, which gives $(W-R)/D$ similarity scores. We compare the similarity scores for frames rotated left, rotated right and unchanged, using the similarity score in rightmost window to predict a left turn, middle window to predict moving forward, and leftmost window to predict right turn. 
Mean squared error over pixel values is used to represent the similarity between two frames, and lower mean squared error indicates higher similarity. 

\subsection{Object Detection} \label{obj_detector}
We utilize a Mask-RCNN model~\cite{he2017mask} from the Detectron2~\cite{wu2019detectron2} package pre-trained on the LVIS dataset~\cite{gupta2019lvis} to detect objects in video frames. The LVIS dataset contains a moderate number of object classes (around 1200 classes), which detects more diverse objects such as ``trash can" compared with MSCOCO (91 classes), and is not too detailed as YOLO9000 which divides car into different classes such as ``hotrod".  
We manually filter out some object classes that appear frequently (i.e., ``bus", ``car(automobile)"), and items that are not helpful for navigation (i.e., ``license plate", ``wheel", ``rearview mirror", ``taillight")

\subsection{Instruction Generation} \label{ins_generation}
Each video in the BDD100K dataset is a 40 seconds driving video. We sample frames with an interval of 1 second. To increase the variance of the trajectory length, we sample a trajectory length between 25 to 40. We predict actions between consecutive frames using our rotation-similarity based action predictor, and further merge predicted successive forward actions together with a maximum of 6 forwards.
Similarly, we merge successive left or right actions together since one turn action might happen across multiple frames. After merging, we generate one sentence for each action and concatenate them as the final instruction. 

%% file: model_training.tex
\section{VLN Model and Training Procedures}

\subsection{Stage 1: Pre-training}

We pre-train the VLN-Transformer model~\cite{zhu-etal-2021-multimodal} 
on the Touchdown dataset~\cite{Chen19:touchdown}, Manh50 dataset~\cite{zhu-etal-2021-multimodal} and BDD100K dataset~\cite{chen2018bdd100k}. Specifically, we use our \methodname{} to generate synthetic instructions for both Manh50 and BDD100K. On Manh50, we use the heading change between nodes on the graph to fill in the actions, while on BDD100K, we use our action predictor to predict the action sequence. 

VLN-Transformer~\cite{zhu-etal-2021-multimodal} is a multi-modal transformer model. 
In this model, words in the instruction are encoded using BERT~\cite{devlin2018bert} and averaged to obtain a sentence embedding for each sentence in the instruction.
View representations of observations are obtained using the fourth-to-last layer features from ResNet~\cite{he2016deep} pre-trained on ImageNet~\cite{deng2009imagenet}, flattened using another convolutional layer.
Sentence and view embeddings are concatenated, passed through cross modal transformer layers and finally concatenated and passed through a fully connected layer to obtain the next action prediction.

We use three proxy tasks to pre-train the model: 
Masked Language Modeling (MLM), Instruction and Trajectory Matching (ITM), and Next Action Prediction (NAP). In Masked Language Modeling, the agent needs to recover masked words given surrounding words and visual information from the full navigation trajectory. In Instruction and Trajectory Matching, we create four negative paths for each positive trajectory and instruction pair, and the agent needs to identify the positive pair. Two of the negative pairs are randomly sampled from the batch, and another two are created by shuffling the sequence of the viewpoints in the trajectory. In this case, the model learns to align both the semantics between instruction and visual observations, and also be aware of the order information of the trajectory. 
In Next Action Prediction, given full navigation instruction and the history observation, the agent predicts the next step action.

\subsection{Stage 2: Fine-tuning.}
In the second stage, we fine-tune the state-of-the-art ORAR model~\cite{schumann2022analyzing} on the navigation task with imitation learning. The ORAR model is an LSTM based encoder-decoder agent. The instructions are encoded with a bi-directional LSTM over word embeddings learned from an embedding layer. The decoder is a two-layer LSTM, where the first layer encodes the trajectory representation and the second layer learns to predict action. The input $i_t = v_t || a_{t-1} || j_t || d_t$ to the first layer LSTM is a concatenation of encoded history visual observations $v_t$, action embeddings at previous step $a_{t-1}$, junction embedding indicates the number of outgoing edges at each step $j_t$, and heading embedding that indicates the heading of each step $d_t$. Soft dot attention is utilized to calculate weighted instruction representation $c_l$ based on trajectory representations, and calculate weighted trajectory representation $c_i$ based on weighted instructions. The policy making LSTM layer predicts the next action given $c_l$, $c_i$, $e_t$, and $h_t^{first}$, where $e_t$ is the time embedding of the current step, and $h_t^{first}$ is the output at step $t$ from the first LSTM layer.

We extract the learned contextualized instruction representation from the pre-trained VLN-Transformer, and use it as the input word embedding to the LSTM based ORAR agent. As the original ORAR paper~\cite{schumann2022analyzing} utilized a randomly initialized embedding layer to learn the word embedding, we additionally compare to using pre-trained BERT-base embeddings~\cite{devlin2018bert} for fair comparison.

%% file: experiments.tex
\section{Experiments}

\subsection{Datasets and Evaluation Metrics}

We evaluate our agent on the Touchdown dataset~\cite{Chen19:touchdown}. Touchdown is set in Manhattan and contains 9,326 instruction-trajectory pairs, with 6,526 examples in the training set, 1,391 examples in the validation set, and 1,409 examples in the test set. 
The dataset contains 29,641 panoramas with 61,319 edges between them. The training set, validation set and test set share the same environment, so that all environments are seen during training.

The Manh50 dataset~\cite{zhu-etal-2021-multimodal} we use during pre-training is extracted from the StreetLearn dataset~\cite{mirowski2019streetlearn} by sampling trajectories in the Manhattan area with length less than 50. 
The Manh50 dataset contains 31k training trajectories, and each training trajectory is paired with one instruction automatically generated by the Google Map API. Prior work transfers the style of these instructions to more natural instructions~\cite{zhu-etal-2021-multimodal}.
We compare this in Sec.~\ref{speaker} to generating synthetic instructions for Manh50 using our instruction rephrasing method, filling templates with ground truth navigation actions.

The driving videos we utilized during pre-training come from the BDD100K dataset~\cite{chen2018bdd100k}. The original BDD100K dataset contains 100K driving videos in multiple cities - New York, Berkeley, San Francisco, and various cities in the Bay Area. We use a subset of 6K videos in the BDD100K training data, where each driving video is captured in a city environment with clear weather during daytime.

We use three metrics to evaluate agents' performance on Touchdown: Task Completion (TC), Shortest Path Distance (SPD), and Success weighted by Edit Distance (SED). Details can be found in Appendix. 

\begin{table}
\centering
\resizebox{1.0\columnwidth}{!}{
\begin{tabular}{p{0.3\columnwidth}|>{\centering\arraybackslash}p{0.08\columnwidth}>{\centering\arraybackslash}p{0.08\columnwidth}>{\centering\arraybackslash}p{0.08\columnwidth}|>{\centering\arraybackslash}p{0.08\columnwidth}>{\centering\arraybackslash}p{0.08\columnwidth}>{\centering\arraybackslash}p{0.08\columnwidth}}
\hline 
\multicolumn{1}{c}{\textbf{Models}} & \multicolumn{3}{c}{\textbf{Validation Set}} & \multicolumn{3}{c}{\textbf{Test Set}}  \\ \hline
 & \textbf{TC$\uparrow$} & \textbf{SPD$\downarrow$} & \textbf{SED$\uparrow$}  & \textbf{TC$\uparrow$} & \textbf{SPD$\downarrow$} & \textbf{SED$\uparrow$} \\ \hline
RConcat~  & 10.6 & 20.4 & 10.3 & 11.8 & 20.4 & 11.5 \\ 
GA & 12.0 & 18.7 & 11.6 & 11.9 & 19.0 & 11.5 \\
ARC-L2S & 19.5 & 17.1 & 19.0 & 16.7 & 18.8 & 16.3 \\
VLN-Trans & 15.0 & 20.3 & 14.7 & 16.2 & 20.8 & 15.7 \\
ORAR & 30.1 &	11.1	& 29.5	& 29.6	& 11.8	& 28.9 \\
PM-VLN\textsuperscript{*} & \textcolor{gray}{33.0} & \textcolor{gray}{23.6} & \textcolor{gray}{29.5} & \textcolor{gray}{33.4} & \textcolor{gray}{23.8} & \textcolor{gray}{29.7} \\
ORAR-BERT &  30.6& 10.3 & 29.9 & 30 & 11.3 & 29 \\ 
Ours &  \textbf{34.5} & \textbf{9.6} &  \textbf{33.5} & \textbf{31.7} & \textbf{11.2} & \textbf{31} \\
\hline
\end{tabular}
}
\caption{Comparison of agent performance on Touchdown dataset on validation set and test set. Agent marked with $*$ has access to ground truth path traces during navigation. See Sec.~\ref{test} for more details.}
\label{table1}
\end{table}

\section{Results and Analysis} 

\subsection{Comparison with SotA Agents} \label{test}
In this section, we compare our agent with previous state-of-the-art approaches. As shown in Table~\ref{table1}, our \methodname{} agent pre-trained on Touchdown and processed BDD100K videos achieves the new state-of-the-art on validation set and test set. 
Note that PM-VLN agent~\cite{armitage2022priority} has access to extra ground truth information - it uses the path trace image of the full navigation trajectory for trajectory planning, where the agent can foresee all future actions implicitly at the start of navigation. 
Though \methodname{} is 1.7\% lower than the PM-VLN agent in task completion on test set, our agent significantly improves the SPD score by 12.6\%, and 1.3\% in SED, demonstrating that our agent follows the instruction better while navigating correctly. ORAR-BERT is the agent that utilizes pre-trained BERT as the embedding input to the ORAR model. \methodname{} improves over ORAR-BERT in task completion by 3.9\% on validation set and 1.7\% on test set, demonstrating that our agent is able to learn better instruction representation through pre-training on both the Touchdown dataset and processed driving videos.

\begin{table}
\centering
\resizebox{1.0\columnwidth}{!}{
\begin{tabular}{p{0.3\columnwidth}|>{\centering\arraybackslash}p{0.08\columnwidth}>{\centering\arraybackslash}p{0.08\columnwidth}>{\centering\arraybackslash}p{0.08\columnwidth}|>{\centering\arraybackslash}p{0.08\columnwidth}>{\centering\arraybackslash}p{0.08\columnwidth}>{\centering\arraybackslash}p{0.08\columnwidth}}
\hline 
\multicolumn{1}{c}{\textbf{Models}} & \multicolumn{3}{c}{\textbf{Validation Set}} & \multicolumn{3}{c}{\textbf{Test Set}}  \\ \hline
 & \textbf{TC$\uparrow$} & \textbf{SPD$\downarrow$} & \textbf{SED$\uparrow$}  & \textbf{TC$\uparrow$} & \textbf{SPD$\downarrow$} & \textbf{SED$\uparrow$} \\ \hline
ORAR-BERT &  30.6& 10.3 & 29.9 & 30 & 11.3 & 29.0 \\ 
+TD  & 31.6 &	10.6 &	30.9 &	31.4 &	11.2 &	30.6 \\ 
+TD+M50-ori  & 32.2 &	10.0	& 31.4	& 30.4	& 10.8	& 29.5\\
+TD+M50-style\textsuperscript{*}  & \textcolor{gray}{31.6}	& \textcolor{gray}{10.0} & \textcolor{gray}{30.7} & \textcolor{gray}{32.7} & \textcolor{gray}{10.5} & \textcolor{gray}{32.0} \\
+TD+M50-our  & 34.3 & 10.7 & \textbf{33.7} & 31.2 & 11.1 & 30.4 \\ 
Ours & \textbf{34.5} & \textbf{9.6} &  33.5 & \textbf{31.7} & \textbf{11.2} & \textbf{31.0} \\
\hline
\end{tabular}
}
\caption{Comparison of agent performance when trained on different datasets on Touchdown validation set and test set. The instruction style transfer model in agent marked with $*$ has access to Touchdown data in validation and test set. See Sec.~\ref{pt} for more details.}
\label{table2}
\end{table}

\begin{table}
\centering
\resizebox{1.0\columnwidth}{!}{
\begin{tabular}{p{0.15\columnwidth}|>{\centering\arraybackslash}p{0.08\columnwidth}>{\centering\arraybackslash}p{0.08\columnwidth}>{\centering\arraybackslash}p{0.08\columnwidth}|>{\centering\arraybackslash}p{0.08\columnwidth}>{\centering\arraybackslash}p{0.08\columnwidth}>{\centering\arraybackslash}p{0.08\columnwidth}}
\hline 
\multicolumn{1}{c}{\textbf{Models}} & \multicolumn{3}{c}{\textbf{Validation Set}} & \multicolumn{3}{c}{\textbf{Test Set}}  \\ \hline
  & \textbf{TC$\uparrow$} & \textbf{SPD$\downarrow$} & \textbf{SED$\uparrow$}  & \textbf{TC$\uparrow$} & \textbf{SPD$\downarrow$} & \textbf{SED$\uparrow$} \\ \hline
Baseline &  30.6& 10.3 & 29.9 & 30.0 & 11.3 & 29.0 \\ 
Ours & \textbf{34.5} & \textbf{9.6} &  \textbf{33.5} & \textbf{31.7} & 11.2 & \textbf{31.0} \\ 
-speaker & 27.1 & 11.7 & 26.4 & 28.2 & 11.7 & 27.6 \\ 
-GPT2 & 31.6 & 10.3 & 30.8 & 29.8 & \textbf{10.3} & 29.1 \\
-action &31.3 & 10.0 & 30.7& 30.7& 11.0 & 30.0 \\ \hline 
\end{tabular}
}
\caption{Ablations of our proposed instruction rephrasing method and action prediction method on Touchdown dataset. The agents are pre-trained on Touchdown dataset and driving videos.}
\label{table3}
\end{table}

\begin{figure}[t]
\begin{center}
\includegraphics[width=1.0\linewidth]{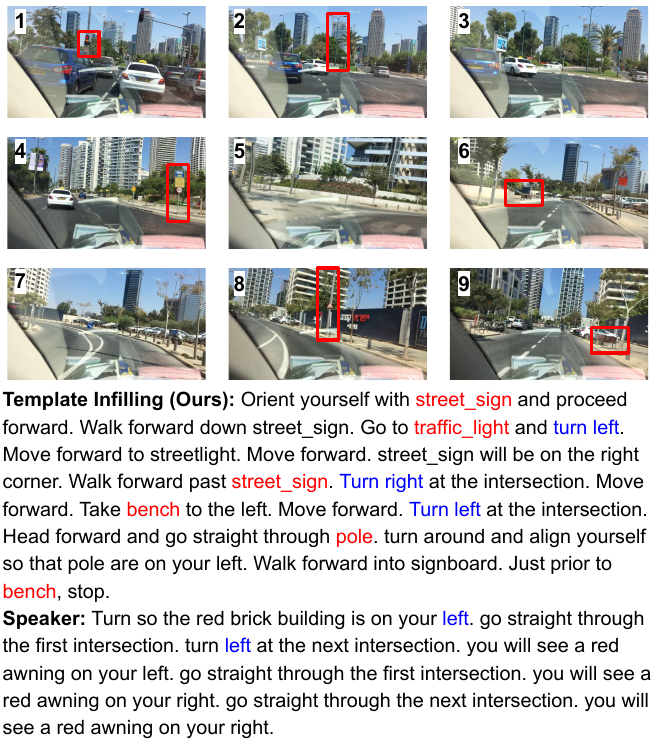}
\end{center}
  \caption{
  Qualitative Analysis of our proposed \methodname{} in generating synthetic instructions for videos in BDD100K dataset. The object entities mentioned in the synthetic instructions generated with our \methodname{} are in red, and their corresponding location is bounded in the red box in the frames. The turn actions are in blue.
}
\label{figure3}
\end{figure}

\subsection{Pre-training with Different Datasets} \label{pt}
We evaluate the agents' performance when pre-trained on different datasets. As shown in Table~\ref{table2}, pre-training on Touchdown only (``+TD") achieves 1.0\% improvement in TC on the validation set, demonstrating that the agent benefits from the in-domain knowledge learned during pre-training. 
Pre-training on Touchdown and Manh50 (``+TD+M50-ori") improves the baseline model by 1.6\% in TC and 0.3\% in SPD on the validation set, suggesting that the agent benefits from learning from more in-domain instruction-trajectory pairs. 
Then we evaluate the influence of instruction quality on performance by comparing a model pre-trained on automatically generated instructions in Manh50 dataset (``+TD+M50-ori"), transferred style instructions provided by \cite{zhu-etal-2021-multimodal} (``+TD+M50-style"), and instructions generated by our method (``+TD+M50-our"). Table~\ref{table2} shows that pre-training with instructions generated by our method achieves the best performance, outperforming the others by more than 2\% in TC on the validation set. Training with transferred style instructions achieves the best performance on test set, which we hypothesize is due to the instruction style transfer model being trained on data from validation and test splits in Touchdown dataset. Finally, pre-training on both Touchdown and driving videos (``Ours") improves the baseline by 3.9\% in TC on the validation set, indicating that the agent benefits from learning from the diverse visual observations and temporal information in the driving videos. 

\subsection{Speaker and Turn Predictor Effectiveness}\label{speaker}
We compare our template infilling method and image rotation similarity based navigation action prediction method with popular deep learning based methods. Specifically, we train a LSTM based speaker model~\cite{tan2019learning} on Touchdown and use it to generate synthetic instructions for driving videos. When combined with actions predicted using our rotation similarity based action predictor, as shown in Table~\ref{table3}, using this learned speaker (``-speaker") decreases the performance by 3.5\% in TC on the validation set, demonstrating that the learned speaker fails to generate good quality instructions for the driving videos. 
Then, we remove the GPT-2 template filtering step during data generation (``-GPT2") and observe a decrease in performance by 2.9\% in TC on validation set, demonstrating that it's crucial to filter out templates with low quality.
We further experiment with utilizing a deep learning based turn predictor trained on Touchdown to predict actions between frames in the videos. The turn predictor is a multi-layer perceptron over visual features encoded with pre-trained ResNet and additional learned CNN layers. When combined with our template infilling method to generate the instructions, as shown in Table~\ref{table3}, the learned turn predictor improves over the baseline by 0.7\% in TC on the validation set, but is 3.2\% lower than using the actions predicted with our rotation-similarity based approach. We also include a human evaluation to show the effectiveness of our action predictor (in Appendix).

\begin{figure}[t]
\begin{center}
\includegraphics[width=1.0\linewidth]{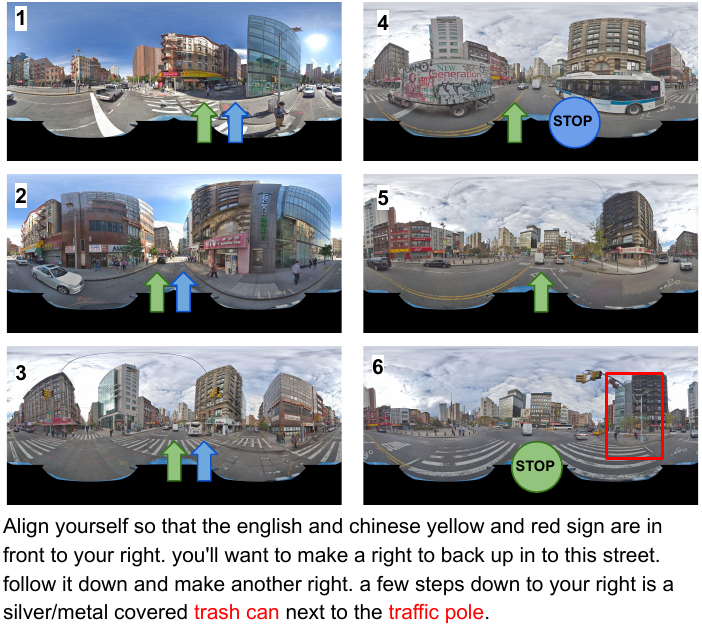}
\end{center}
  \caption{
  Qualitative Analysis of our proposed \methodname{} in learning richer visual objects to help decision making during navigation. Symbols in green are the actions made by our method, and symbols in blue are the actions made by the baseline method. 
}
\label{figure4}
\end{figure}

\subsection{Qualitative Analysis of Synthetic Instructions} \label{quality}
We show a qualitative example of our generated synthetic instructions on the BDD100K dataset. As shown in Figure~\ref{figure3}, given the driving video shown here, \methodname{} is able to generate long instructions with rich entity mentions. In the video, the car first turns left at the traffic light, then turn right at the branch road, and finally the cars follow the road to drive left. The synthetic instructions mention four different objects (street sign, traffic light, bench, pole) which appear in the videos. Furthermore, our rotation based action prediction method successfully predicts all the three turns made in the instruction. In comparison, the instruction generated by the speaker method is repetitive and mentions objects like ``awning" that do not appear in the views.

\subsection{Case Study}
We show a qualitative example to compare the navigation trajectory predicted by \methodname{} and the baseline. As shown in Figure~\ref{figure4}, \methodname{} successfully identifies the ``trash can" and ``traffic pole" in the view and stops correctly, whereas the baseline fails and stops earlier. This suggests that diverse visual data seen in the videos helps the agent learn the semantics of objects better.

%% file: conclusion.tex
\section{Conclusion}
In this paper, we propose \methodname{} - a method to process driving videos to obtain augmented data for outdoor Vision-and-Language Navigation by utilizing intuitive classical techniques for navigation action prediction and instruction generation, which when combined with recent deep learning models for VLN obtains a new state-of-the-art performance on the Touchdown dataset.
We propose an image rotation similarity based method to predict navigation actions between consecutive video frames and a template infilling based method for instruction generation that does not require any human annotation.
Our method can more generally be applied to preprocess any video dataset for pre-training any VLN model. 
Our experiments on the Touchdown dataset show that pre-training the agent with our generated synthetic navigation data helps the agent learn contextualized language representations that ground better to the navigation environments, which greatly improves the downstream navigation performance. 
\methodname{} achieves the new state-of-the-art performance on Touchdown dataset, and provides a good starting point for future work that aims to exploit the rich visual and temporal information in videos for data augmentation in VLN tasks.

%% file: Supplementary.tex
\section{Appendix Overview}
In this supplementary, we provide the following:
\begin{itemize}
    \item Detailed analysis on whether the agent could learn to localize the more relevant parts in the panorama in Sec.~\ref{localize}. 
    \item We further include two more examples showing that our agent is able to learn richer visual objects to help decision making during navigation in Sec.~\ref{examples}. 
    \item Then, we show the our agents' generalization performance on unseen environments on re-splitted Touchdown dataset in Sec.~\ref{generalization}. 
    \item Moreover, we include human evaluation of the actions predicted with our rotation similarity based action predictor in Sec.~\ref{human_eval}. 
    \item Lastly, we provide implementation details and evaluation metrics in Sec.~\ref{implementation} and Sec.~\ref{evaluation}.
\end{itemize}

\section{Analysis: Learning to Attend Over Panorama} \label{localize}
In this section, we evaluate whether an agent could locate the important parts in the observation when given the full panorama. The ORAR~\cite{schumann2022analyzing} takes the middle parts of the panorama based on heading as the input to the agent, so that the agent could learn to attend over objects in a smaller area. Specifically, they split the panorama into 8 views, and each view is encoded with ResNet18~\cite{he2016deep} pre-trained on ImageNet~\cite{deng2009imagenet}. They concatenate the fourth-to-last layer outputs from the pre-trained ResNet18 ($128\times100\times464$) and pick the middle 100 dimensions ($128\times100\times100$) (around 25\% of the full panorama) as the panorama representation. We further experiment with using a larger portion of the panorama as input to test agents' ability to learn the alignment between the instruction and the observation. Specifically, we split the panorama observation into 36 discretized views, and each view feature is extracted from the last layer of ResNet18~\cite{he2016deep} pre-trained on ImageNet~\cite{deng2009imagenet}. Given 36 encoded discretized views $\{v_{ji}\}_{i=0}^{36}$ at navigation step $j$, we represent the panorama observation $p_j$ as the attended 36 views based on trajectory representation $s_{j-1}$:

\begin{align}
    \alpha_i &= softmax_i(v_{ji}^T W_F s_{j-1}) \\ 
    p_j &= \sum_{i=0}^{36}{\alpha_i v_{ji}}
\end{align}

$W_F$ is a learable parameter. The trajectory representation $s_{j-1}$ is the output of the first LSTM layer in the decoder, where $s_{j-1} = h_j^{first}$ in the main paper.

\begin{table}
\centering
\begin{tabular}{p{0.17\columnwidth}|>{\centering\arraybackslash}p{0.08\columnwidth}>{\centering\arraybackslash}p{0.08\columnwidth}>{\centering\arraybackslash}p{0.08\columnwidth}|>{\centering\arraybackslash}p{0.08\columnwidth}>{\centering\arraybackslash}p{0.08\columnwidth}>{\centering\arraybackslash}p{0.08\columnwidth}}
\hline 
\multicolumn{1}{c}{\textbf{Models}} & \multicolumn{3}{c}{\textbf{Validation Set}} & \multicolumn{3}{c}{\textbf{Test Set}}  \\ \hline
  & \textbf{TC$\uparrow$} & \textbf{SPD$\downarrow$} & \textbf{SED$\uparrow$}  & \textbf{TC$\uparrow$} & \textbf{SPD$\downarrow$} & \textbf{SED$\uparrow$} \\ \hline
ORAR &  30.1 & 11.1 & 29.5 & 29.6 & 11.8 & 28.9 \\
36-views &  19.3 & 23.2 & 18.8 & 17.9 & 24.2 & 17.5 \\ 
+action &  28.4 & 11.3 & 27.8 & 27.3 & 11.3 & 26.7 \\ 
+drop &  31.0 & 10.5 & 30.4 & 29.6 & 10.9 & 28.9  \\ \hline 
\end{tabular}
\caption{Analysis of how well the agent learns to attend over full panorama.}
\label{appendix_table1}
\end{table}

As shown in Table~\ref{appendix_table1}, directly using attended 36 views as the panorama representation (``36-views") results in a significant drop in performance, decreasing the task completion rate by 10.8\% on the validation set. This demonstrates that only using the middle parts in the full panorama greatly helps the agent to localize the relevant navigation information in the panorama, and the agent could not learn to pick the more important views given the instruction. 

We further enhance the ORAR~\cite{schumann2022analyzing} model's action space with observation features. Specifically, for the four possible actions ``forward", ``left", ``right" and ``stop", we represent each action with the view representation at their corresponding heading. The agent could benefit from the visual information underlying the action representation to make better decisions. As shown in Table~\ref{appendix_table1}, enhancing the action representation with the visual information (``+action") greatly improves the performance by 9.1\% in task completion. This demonstrates that the discretized view representation in the action space helps the agent to make the correct navigation action, especially when the agent takes the full panorama as input and fails to localize the navigation direction. 

Lastly, we try to drop half of the observations in the panorama where the agent is not facing, and calculate the panorama representation only with the 18 front views. As shown in Table~\ref{appendix_table1}, dropping half of the views (``+drop") further improves the task completion by 2.6\%, and achieves similar performance as the baseline ORAR~\cite{schumann2022analyzing}, which utilizes only 25\% of the panorama observation during navigation. This demonstrates that the agent could not learn to drop out irrelevant views based on instructions, and needs explicit supervision on which views in the panorama to look at.

\section{Case Study: Demonstrating Improvements in Visual Understanding} \label{examples}

\begin{figure}[t]
\begin{center}
\includegraphics[width=1.0\linewidth]{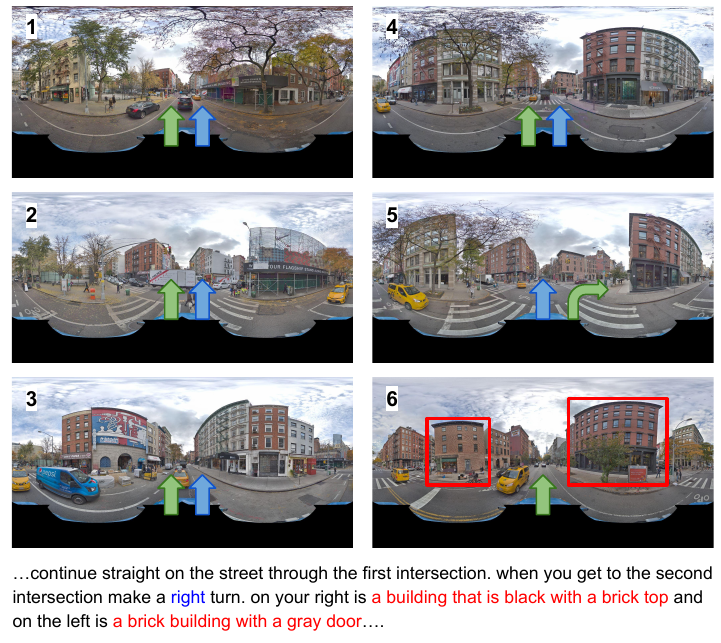}
\end{center}
  \caption{
  Qualitative Analysis of our proposed \methodname{} in learning richer visual objects to help decision making during navigation. Symbols in green are the actions made by our method, and symbols in blue are the actions made by the baseline method. 
}
\label{appendix_figure1}
\end{figure}

\begin{figure}[t]
\begin{center}
\includegraphics[width=1.0\linewidth]{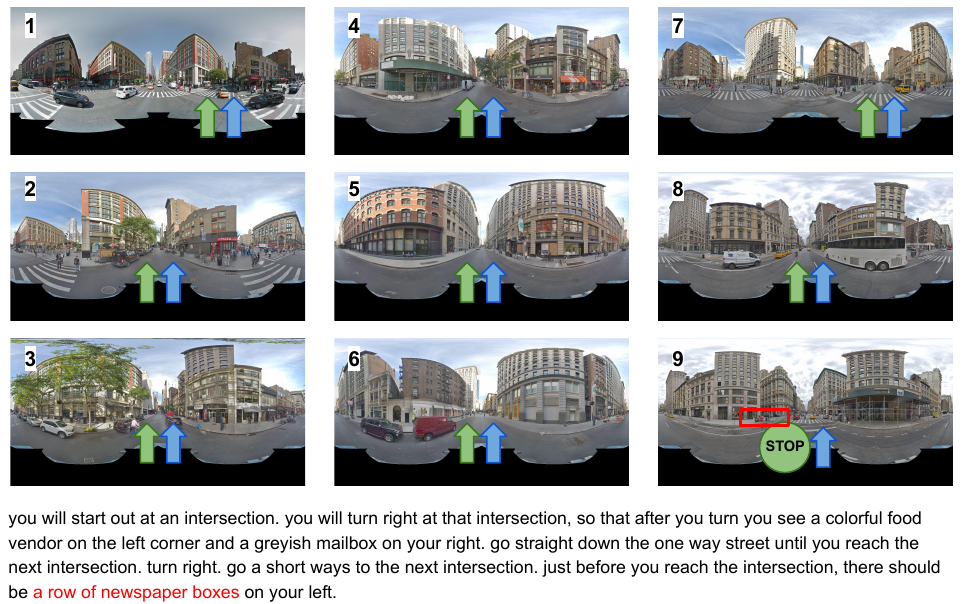}
\end{center}
  \caption{
  Qualitative Analysis of our proposed \methodname{} in learning richer visual objects to help decision making during navigation. Symbols in green are the actions made by our method, and symbols in blue are the actions made by the baseline method. 
}
\label{appendix_figure2}
\end{figure}

In this section, we show more examples where our agent is able to learn richer visual objects to help decision-making during navigation. As shown in Figure~\ref{appendix_figure1}, the agent learned with our \methodname{} is able to turn right at the correct intersection by identifying the ``building that is black with a brick top" and ``a brick building with a gray door" while the baseline agent fails and moves forward through the intersection. Besides, we show in Figure~\ref{appendix_figure2} that our agent is able to stop correctly beside the ``newspaper boxes" where the baseline agent moves past the boxes and stops at a wrong location.

\begin{table}
\centering
\resizebox{1.0\columnwidth}{!}{
\begin{tabular}{p{0.3\columnwidth}|>{\centering\arraybackslash}p{0.08\columnwidth}>{\centering\arraybackslash}p{0.08\columnwidth}>{\centering\arraybackslash}p{0.08\columnwidth}|>{\centering\arraybackslash}p{0.08\columnwidth}>{\centering\arraybackslash}p{0.08\columnwidth}>{\centering\arraybackslash}p{0.08\columnwidth}}
\hline 
\multicolumn{1}{c}{\textbf{Models}} & \multicolumn{3}{c}{\textbf{Validation Set}} & \multicolumn{3}{c}{\textbf{Test Set}}  \\ \hline
  & \textbf{TC$\uparrow$} & \textbf{SPD$\downarrow$} & \textbf{SED$\uparrow$}  & \textbf{TC$\uparrow$} & \textbf{SPD$\downarrow$} & \textbf{SED$\uparrow$} \\ \hline
ORAR & 15.4 & - & - & 14.9 & - & -   \\
ORAR-BERT & 17.5 & 20.6 & 16.8 & 15.7 & 21.6 & 15 \\ 
+TD & 18.4 &	20.2	& 17.7	& 14.9 &	20.7	& 14.3 \\
+TD+M50-ori & 18.8 &	\textbf{19.8}	& 18.2	& \textbf{16.4}	& \textbf{20.1}	& \textbf{15.8} \\ 
+TD+M50-style & 18.4 &	20.1 &	17.8 & 	16.1 &	20.1	& 15.5\\ 
+TD+M50-our & \textbf{19.9}	& 20.0 &	\textbf{19.1}	& 15.3	& 20.1	& 14.6\\ 
Ours & 18.2	& 20.2	& 17.5	& 16.3	& 21.2	& 15.7 \\ \hline 
\end{tabular}
}
\caption{Performance of our method on unseen split in re-splitted Touchdown dataset.}
\label{appendix_table2}
\end{table}

\section{Analysis: Generalization to Unseen Environments} \label{generalization}
In this section, we analyze the generalization performance of our agent to unseen environments. Specifically, we follow~\cite{schumann2022analyzing} to resplit the Touchdown dataset into training set, unseen validation set, and unseen test set, where the validation set and test set instances are from the unseen environments. As shown in Table~\ref{appendix_table2}, pre-training on both the Touchdown dataset and the video datasets significantly improves agents' task completion rate by 2.5\% on test unseen set, demonstrating that our method helps generalization to unseen environments. We also notice that when the model is pre-trained on both the Touchdown dataset and the Manh50 dataset, the task completion improves the baseline by more than 3\% in task completion on validation unseen set. This is because the Manh50 dataset contains environments in the Manhanttan area, which might have environments in the unseen validation set. Among them, pre-training on the Touchdown dataset and Manh50 dataset with the instructions generated by our method achieves the largest improvement in task completion, demonstrating the effectiveness of our proposed instruction infilling method.

\section{Human Evaluation: Accuracy of Action Predictor} \label{human_eval}
We include a brief human evaluation for the accuracy of our action predictor. We sample 50 videos from BDD100K training set, and sample two consecutive frames from each of the videos. We evaluate our rotation similarity based turn predictor's performance on these 50 examples with two human annotators. Predicting the turn between two consecutive frames is not a simple task, where the inter-annotator agreement between two annotators is only 0.493, demonstrating moderate agreement between two annotators. Our rotation-similarity based turn predictor is able to predict 76\% of the actions that correctly match at least one of the annotators, while the predicted turn predictor could only match 74\% of the actions. More importantly, among them, our rotation-similarity based turn predictor successfully predicts 33\% of the turns correctly, while the learned turn predictor \textbf{fails on all the turns}. This demonstrates the effectiveness of our rotation based turn predictor, and suggests that how to predict turn movements in city environment in a zero-shot way can be an important task for future research.

\section{Implementation Details} \label{implementation}
In the pre-training stage, the model is trained on the three proxy tasks with equal weight for 200k iterations with a batch size of 64. 
In the fine-tuning stage, we follow the hyperparameters used in the ORAR model, where we train the agent for 300 epochs with a batch size of 64. AdamW is used to optimize the model with a learning rate of 5e-4 and weight decay of 1e-2. In image rotation similarity based navigation action prediction, we use a window size of 80 degrees, and a shift angle of 60 degrees.

\section{Evaluation Metrics} \label{evaluation}
We use three metrics to evaluate agents' performance on Touchdown: (1) Task Completion (TC): if the agent stops at the target viewpoint or any of its neighbouring viewpoints, we consider the navigation task as complete. (2) Shortest Path Distance (SPD): it measures the shortest distance between the agents' final position and the goal position in the environment. Lower SPD score indicates the agent is closer to the target location. (3) Success weighted by Edit Distance (SED): it weights the task completion with the Levenshtein edit distance between the predicted trajectory and the ground truth trajectory. This metric penalizes agents that succeed by exploring the environment rather than following instructions.

%% file: aaai24.bbl
\begin{thebibliography}{68}
\providecommand{\natexlab}[1]{#1}

\bibitem[{Abu-El-Haija et~al.(2016)Abu-El-Haija, Kothari, Lee, Natsev, Toderici, Varadarajan, and Vijayanarasimhan}]{abu2016youtube}
Abu-El-Haija, S.; Kothari, N.; Lee, J.; Natsev, P.; Toderici, G.; Varadarajan, B.; and Vijayanarasimhan, S. 2016.
\newblock Youtube-8m: A large-scale video classification benchmark.
\newblock \emph{arXiv preprint arXiv:1609.08675}.

\bibitem[{Akbik, Blythe, and Vollgraf(2018)}]{akbik2018coling}
Akbik, A.; Blythe, D.; and Vollgraf, R. 2018.
\newblock Contextual String Embeddings for Sequence Labeling.
\newblock In \emph{{COLING} 2018, 27th International Conference on Computational Linguistics}, 1638--1649.

\bibitem[{Anderson et~al.(2018)Anderson, Wu, Teney, Bruce, Johnson, S{\"u}nderhauf, Reid, Gould, and van~den Hengel}]{anderson2018room2room}
Anderson, P.; Wu, Q.; Teney, D.; Bruce, J.; Johnson, M.; S{\"u}nderhauf, N.; Reid, I.; Gould, S.; and van~den Hengel, A. 2018.
\newblock Vision-and-language navigation: Interpreting visually-grounded navigation instructions in real environments.
\newblock In \emph{Proceedings of the IEEE Conference on Computer Vision and Pattern Recognition}, 3674--3683.

\bibitem[{Armitage, Impett, and Sennrich(2022)}]{armitage2022priority}
Armitage, J.; Impett, L.; and Sennrich, R. 2022.
\newblock A Priority Map for Vision-and-Language Navigation with Trajectory Plans and Feature-Location Cues.
\newblock \emph{arXiv preprint arXiv:2207.11717}.

\bibitem[{Baker et~al.(2022)Baker, Akkaya, Zhokhov, Huizinga, Tang, Ecoffet, Houghton, Sampedro, and Clune}]{baker2022video}
Baker, B.; Akkaya, I.; Zhokhov, P.; Huizinga, J.; Tang, J.; Ecoffet, A.; Houghton, B.; Sampedro, R.; and Clune, J. 2022.
\newblock Video pretraining (vpt): Learning to act by watching unlabeled online videos.
\newblock \emph{arXiv preprint arXiv:2206.11795}.

\bibitem[{Berg et~al.(2020)Berg, Bayazit, Mathew, Rotter-Aboyoun, Pavlick, and Tellex}]{berg2020grounding}
Berg, M.; Bayazit, D.; Mathew, R.; Rotter-Aboyoun, A.; Pavlick, E.; and Tellex, S. 2020.
\newblock Grounding Language to Landmarks in Arbitrary Outdoor Environments.
\newblock In \emph{2020 IEEE International Conference on Robotics and Automation (ICRA)}, 208--215. IEEE.

\bibitem[{Blukis et~al.(2019)Blukis, Terme, Niklasson, Knepper, and Artzi}]{blukis2020learning}
Blukis, V.; Terme, Y.; Niklasson, E.; Knepper, R.~A.; and Artzi, Y. 2019.
\newblock Learning to Map Natural Language Instructions to Physical Quadcopter Control using Simulated Flight.
\newblock In \emph{Conference on Robot Learning}, 1415--1438.

\bibitem[{Brahmbhatt and Hays(2017)}]{brahmbhatt2017deepnav}
Brahmbhatt, S.; and Hays, J. 2017.
\newblock Deepnav: Learning to navigate large cities.
\newblock In \emph{Proceedings of the IEEE Conference on Computer Vision and Pattern Recognition}, 5193--5202.

\bibitem[{Chang et~al.(2017)Chang, Dai, Funkhouser, Halber, Niessner, Savva, Song, Zeng, and Zhang}]{chang2017matterport3d}
Chang, A.; Dai, A.; Funkhouser, T.; Halber, M.; Niessner, M.; Savva, M.; Song, S.; Zeng, A.; and Zhang, Y. 2017.
\newblock Matterport3d: Learning from rgb-d data in indoor environments.
\newblock \emph{arXiv preprint arXiv:1709.06158}.

\bibitem[{Chen et~al.(2019)Chen, Suhr, Misra, Snavely, and Artzi}]{Chen19:touchdown}
Chen, H.; Suhr, A.; Misra, D.; Snavely, N.; and Artzi, Y. 2019.
\newblock Touchdown: Natural Language Navigation and Spatial Reasoning in Visual Street Environments.
\newblock In \emph{Conference on Computer Vision and Pattern Recognition}.

\bibitem[{Chen et~al.(2022{\natexlab{a}})Chen, Ji, Lin, Zeng, Li, Tan, and Gan}]{chen2022weakly}
Chen, P.; Ji, D.; Lin, K.; Zeng, R.; Li, T.~H.; Tan, M.; and Gan, C. 2022{\natexlab{a}}.
\newblock Weakly-supervised multi-granularity map learning for vision-and-language navigation.
\newblock \emph{arXiv preprint arXiv:2210.07506}.

\bibitem[{Chen et~al.(2021)Chen, Guhur, Schmid, and Laptev}]{chen2021history}
Chen, S.; Guhur, P.-L.; Schmid, C.; and Laptev, I. 2021.
\newblock History aware multimodal transformer for vision-and-language navigation.
\newblock \emph{Advances in Neural Information Processing Systems}, 34: 5834--5847.

\bibitem[{Chen et~al.(2022{\natexlab{b}})Chen, Guhur, Tapaswi, Schmid, and Laptev}]{chen2022learning}
Chen, S.; Guhur, P.-L.; Tapaswi, M.; Schmid, C.; and Laptev, I. 2022{\natexlab{b}}.
\newblock Learning from Unlabeled 3D Environments for Vision-and-Language Navigation.
\newblock In \emph{European Conference on Computer Vision}, 638--655. Springer.

\bibitem[{Chen et~al.(2018)Chen, Darrell, Yu, and Chen}]{chen2018bdd100k}
Chen, X. W. W. X.~Y.; Darrell, F. L. V. M.~T.; Yu, F.; and Chen, H. 2018.
\newblock Bdd100k: A diverse driving dataset for heterogeneous multitask learning.
\newblock \emph{arXiv preprint arXiv: 1805.04687}.

\bibitem[{Deng et~al.(2009)Deng, Dong, Socher, Li, Li, and Fei-Fei}]{deng2009imagenet}
Deng, J.; Dong, W.; Socher, R.; Li, L.-J.; Li, K.; and Fei-Fei, L. 2009.
\newblock Imagenet: A large-scale hierarchical image database.
\newblock In \emph{2009 IEEE conference on computer vision and pattern recognition}, 248--255. Ieee.

\bibitem[{Devlin et~al.(2018)Devlin, Chang, Lee, and Toutanova}]{devlin2018bert}
Devlin, J.; Chang, M.-W.; Lee, K.; and Toutanova, K. 2018.
\newblock Bert: Pre-training of deep bidirectional transformers for language understanding.
\newblock \emph{arXiv preprint arXiv:1810.04805}.

\bibitem[{Dou and Peng(2022)}]{dou2022foam}
Dou, Z.-Y.; and Peng, N. 2022.
\newblock FOAM: A Follower-aware Speaker Model For Vision-and-Language Navigation.
\newblock \emph{arXiv preprint arXiv:2206.04294}.

\bibitem[{Fabian Caba~Heilbron and Niebles(2015)}]{caba2015activitynet}
Fabian Caba~Heilbron, B.~G., Victor~Escorcia; and Niebles, J.~C. 2015.
\newblock ActivityNet: A Large-Scale Video Benchmark for Human Activity Understanding.
\newblock In \emph{Proceedings of the IEEE Conference on Computer Vision and Pattern Recognition}, 961--970.

\bibitem[{Fried et~al.(2018)Fried, Hu, Cirik, Rohrbach, Andreas, Morency, Berg-Kirkpatrick, Saenko, Klein, and Darrell}]{fried2018speaker}
Fried, D.; Hu, R.; Cirik, V.; Rohrbach, A.; Andreas, J.; Morency, L.-P.; Berg-Kirkpatrick, T.; Saenko, K.; Klein, D.; and Darrell, T. 2018.
\newblock Speaker-follower models for vision-and-language navigation.
\newblock \emph{Advances in Neural Information Processing Systems}, 31.

\bibitem[{Fu et~al.(2020)Fu, Wang, Peterson, Grafton, Eckstein, and Wang}]{fu2020counterfactual}
Fu, T.-J.; Wang, X.~E.; Peterson, M.~F.; Grafton, S.~T.; Eckstein, M.~P.; and Wang, W.~Y. 2020.
\newblock Counterfactual vision-and-language navigation via adversarial path sampler.
\newblock In \emph{Computer Vision--ECCV 2020: 16th European Conference, Glasgow, UK, August 23--28, 2020, Proceedings, Part VI 16}, 71--86. Springer.

\bibitem[{Georgakis et~al.(2022)Georgakis, Schmeckpeper, Wanchoo, Dan, Miltsakaki, Roth, and Daniilidis}]{georgakis2022cross}
Georgakis, G.; Schmeckpeper, K.; Wanchoo, K.; Dan, S.; Miltsakaki, E.; Roth, D.; and Daniilidis, K. 2022.
\newblock Cross-modal map learning for vision and language navigation.
\newblock In \emph{Proceedings of the IEEE/CVF Conference on Computer Vision and Pattern Recognition}, 15460--15470.

\bibitem[{Goyal et~al.(2017)Goyal, Ebrahimi~Kahou, Michalski, Materzynska, Westphal, Kim, Haenel, Fruend, Yianilos, Mueller-Freitag et~al.}]{goyal2017something}
Goyal, R.; Ebrahimi~Kahou, S.; Michalski, V.; Materzynska, J.; Westphal, S.; Kim, H.; Haenel, V.; Fruend, I.; Yianilos, P.; Mueller-Freitag, M.; et~al. 2017.
\newblock The" something something" video database for learning and evaluating visual common sense.
\newblock In \emph{Proceedings of the IEEE international conference on computer vision}, 5842--5850.

\bibitem[{Grauman et~al.(2022)Grauman, Westbury, Byrne, Chavis, Furnari, Girdhar, Hamburger, Jiang, Liu, Liu et~al.}]{grauman2022ego4d}
Grauman, K.; Westbury, A.; Byrne, E.; Chavis, Z.; Furnari, A.; Girdhar, R.; Hamburger, J.; Jiang, H.; Liu, M.; Liu, X.; et~al. 2022.
\newblock Ego4d: Around the world in 3,000 hours of egocentric video.
\newblock In \emph{Proceedings of the IEEE/CVF Conference on Computer Vision and Pattern Recognition}, 18995--19012.

\bibitem[{Guhur et~al.(2021)Guhur, Tapaswi, Chen, Laptev, and Schmid}]{guhur2021airbert}
Guhur, P.-L.; Tapaswi, M.; Chen, S.; Laptev, I.; and Schmid, C. 2021.
\newblock Airbert: In-domain pretraining for vision-and-language navigation.
\newblock In \emph{Proceedings of the IEEE/CVF International Conference on Computer Vision}, 1634--1643.

\bibitem[{Gupta, Dollar, and Girshick(2019)}]{gupta2019lvis}
Gupta, A.; Dollar, P.; and Girshick, R. 2019.
\newblock Lvis: A dataset for large vocabulary instance segmentation.
\newblock In \emph{Proceedings of the IEEE/CVF conference on computer vision and pattern recognition}, 5356--5364.

\bibitem[{Hao et~al.(2020)Hao, Li, Li, Carin, and Gao}]{hao2020towards}
Hao, W.; Li, C.; Li, X.; Carin, L.; and Gao, J. 2020.
\newblock Towards learning a generic agent for vision-and-language navigation via pre-training.
\newblock In \emph{Proceedings of the IEEE/CVF Conference on Computer Vision and Pattern Recognition}, 13137--13146.

\bibitem[{He et~al.(2017)He, Gkioxari, Doll{\'a}r, and Girshick}]{he2017mask}
He, K.; Gkioxari, G.; Doll{\'a}r, P.; and Girshick, R. 2017.
\newblock Mask r-cnn.
\newblock In \emph{Proceedings of the IEEE international conference on computer vision}, 2961--2969.

\bibitem[{He et~al.(2021)He, Huang, Wu, Yang, An, Sima, and Wang}]{he2021landmark}
He, K.; Huang, Y.; Wu, Q.; Yang, J.; An, D.; Sima, S.; and Wang, L. 2021.
\newblock Landmark-RxR: Solving Vision-and-Language Navigation with Fine-Grained Alignment Supervision.
\newblock \emph{Advances in Neural Information Processing Systems}, 34: 652--663.

\bibitem[{He et~al.(2016)He, Zhang, Ren, and Sun}]{he2016deep}
He, K.; Zhang, X.; Ren, S.; and Sun, J. 2016.
\newblock Deep residual learning for image recognition.
\newblock In \emph{Proceedings of the IEEE conference on computer vision and pattern recognition}, 770--778.

\bibitem[{Hermann et~al.(2017)Hermann, Hill, Green, Wang, Faulkner, Soyer, Szepesvari, Czarnecki, Jaderberg, Teplyashin et~al.}]{hermann2017grounded}
Hermann, K.~M.; Hill, F.; Green, S.; Wang, F.; Faulkner, R.; Soyer, H.; Szepesvari, D.; Czarnecki, W.~M.; Jaderberg, M.; Teplyashin, D.; et~al. 2017.
\newblock Grounded language learning in a simulated 3d world.
\newblock \emph{arXiv preprint arXiv:1706.06551}.

\bibitem[{Hermann et~al.(2020)Hermann, Malinowski, Mirowski, Banki-Horvath, Anderson, and Hadsell}]{hermann2019learning}
Hermann, K.~M.; Malinowski, M.; Mirowski, P.; Banki-Horvath, A.; Anderson, K.; and Hadsell, R. 2020.
\newblock Learning to follow directions in street view.
\newblock \emph{Thirty-Fourth AAAI Conference on Artificial Intelligence}.

\bibitem[{Jang et~al.(2017)Jang, Song, Yu, Kim, and Kim}]{jang2017tgif}
Jang, Y.; Song, Y.; Yu, Y.; Kim, Y.; and Kim, G. 2017.
\newblock Tgif-qa: Toward spatio-temporal reasoning in visual question answering.
\newblock In \emph{Proceedings of the IEEE conference on computer vision and pattern recognition}, 2758--2766.

\bibitem[{Kay et~al.(2017)Kay, Carreira, Simonyan, Zhang, Hillier, Vijayanarasimhan, Viola, Green, Back, Natsev et~al.}]{kay2017kinetics}
Kay, W.; Carreira, J.; Simonyan, K.; Zhang, B.; Hillier, C.; Vijayanarasimhan, S.; Viola, F.; Green, T.; Back, T.; Natsev, P.; et~al. 2017.
\newblock The kinetics human action video dataset.
\newblock \emph{arXiv preprint arXiv:1705.06950}.

\bibitem[{Kim, Li, and Bansal(2021)}]{kim2021ndh}
Kim, H.; Li, J.; and Bansal, M. 2021.
\newblock Ndh-full: Learning and evaluating navigational agents on full-length dialogue.
\newblock In \emph{Proceedings of the 2021 Conference on Empirical Methods in Natural Language Processing}, 6432--6442.

\bibitem[{Krishna et~al.(2017)Krishna, Hata, Ren, Fei-Fei, and Carlos~Niebles}]{krishna2017dense}
Krishna, R.; Hata, K.; Ren, F.; Fei-Fei, L.; and Carlos~Niebles, J. 2017.
\newblock Dense-captioning events in videos.
\newblock In \emph{Proceedings of the IEEE international conference on computer vision}, 706--715.

\bibitem[{Ku et~al.(2020)Ku, Anderson, Patel, Ie, and Baldridge}]{ku2020room}
Ku, A.; Anderson, P.; Patel, R.; Ie, E.; and Baldridge, J. 2020.
\newblock Room-Across-Room: Multilingual Vision-and-Language Navigation with Dense Spatiotemporal Grounding.
\newblock In \emph{Proceedings of the 2020 Conference on Empirical Methods in Natural Language Processing (EMNLP)}, 4392--4412.

\bibitem[{Lei et~al.(2018)Lei, Yu, Bansal, and Berg}]{lei2018tvqa}
Lei, J.; Yu, L.; Bansal, M.; and Berg, T.~L. 2018.
\newblock Tvqa: Localized, compositional video question answering.
\newblock \emph{arXiv preprint arXiv:1809.01696}.

\bibitem[{Li, Tan, and Bansal(2022)}]{li2022envedit}
Li, J.; Tan, H.; and Bansal, M. 2022.
\newblock EnvEdit: Environment Editing for Vision-and-Language Navigation.
\newblock In \emph{Proceedings of the IEEE/CVF Conference on Computer Vision and Pattern Recognition}, 15407--15417.

\bibitem[{Lin et~al.(2022{\natexlab{a}})Lin, Zhu, Chen, Liang, Liu, and Liang}]{lin2022adapt}
Lin, B.; Zhu, Y.; Chen, Z.; Liang, X.; Liu, J.; and Liang, X. 2022{\natexlab{a}}.
\newblock ADAPT: Vision-Language Navigation with Modality-Aligned Action Prompts.
\newblock In \emph{Proceedings of the IEEE/CVF Conference on Computer Vision and Pattern Recognition}, 15396--15406.

\bibitem[{Lin et~al.(2022{\natexlab{b}})Lin, Jiang, Cai, Qu, Haffari, and Yuan}]{lin2022multimodal}
Lin, C.; Jiang, Y.; Cai, J.; Qu, L.; Haffari, G.; and Yuan, Z. 2022{\natexlab{b}}.
\newblock Multimodal transformer with variable-length memory for vision-and-language navigation.
\newblock In \emph{European Conference on Computer Vision}, 380--397. Springer.

\bibitem[{Liu et~al.(2021)Liu, Zhu, Chang, Liang, Ge, and Shen}]{liu2021vision}
Liu, C.; Zhu, F.; Chang, X.; Liang, X.; Ge, Z.; and Shen, Y.-D. 2021.
\newblock Vision-language navigation with random environmental mixup.
\newblock In \emph{Proceedings of the IEEE/CVF International Conference on Computer Vision}, 1644--1654.

\bibitem[{Majumdar et~al.(2020)Majumdar, Shrivastava, Lee, Anderson, Parikh, and Batra}]{majumdar2020improving}
Majumdar, A.; Shrivastava, A.; Lee, S.; Anderson, P.; Parikh, D.; and Batra, D. 2020.
\newblock Improving vision-and-language navigation with image-text pairs from the web.
\newblock In \emph{European Conference on Computer Vision}, 259--274. Springer.

\bibitem[{Mehta et~al.(2020)Mehta, Artzi, Baldridge, Ie, and Mirowski}]{mehta2020retouchdown}
Mehta, H.; Artzi, Y.; Baldridge, J.; Ie, E.; and Mirowski, P. 2020.
\newblock Retouchdown: Adding touchdown to streetlearn as a shareable resource for language grounding tasks in street view.
\newblock \emph{arXiv preprint arXiv:2001.03671}.

\bibitem[{Mirowski et~al.(2019)Mirowski, Banki-Horvath, Anderson, Teplyashin, Hermann, Malinowski, Grimes, Simonyan, Kavukcuoglu, Zisserman et~al.}]{mirowski2019streetlearn}
Mirowski, P.; Banki-Horvath, A.; Anderson, K.; Teplyashin, D.; Hermann, K.~M.; Malinowski, M.; Grimes, M.~K.; Simonyan, K.; Kavukcuoglu, K.; Zisserman, A.; et~al. 2019.
\newblock The streetlearn environment and dataset.
\newblock \emph{arXiv preprint arXiv:1903.01292}.

\bibitem[{Mirowski et~al.(2018)Mirowski, Grimes, Malinowski, Hermann, Anderson, Teplyashin, Simonyan, Zisserman, Hadsell et~al.}]{mirowski2018learning}
Mirowski, P.; Grimes, M.; Malinowski, M.; Hermann, K.~M.; Anderson, K.; Teplyashin, D.; Simonyan, K.; Zisserman, A.; Hadsell, R.; et~al. 2018.
\newblock Learning to navigate in cities without a map.
\newblock In \emph{Advances in Neural Information Processing Systems}, 2419--2430.

\bibitem[{Misra et~al.(2018)Misra, Bennett, Blukis, Niklasson, Shatkhin, and Artzi}]{misra2018mapping}
Misra, D.; Bennett, A.; Blukis, V.; Niklasson, E.; Shatkhin, M.; and Artzi, Y. 2018.
\newblock Mapping Instructions to Actions in 3D Environments with Visual Goal Prediction.
\newblock In \emph{Proceedings of the 2018 Conference on Empirical Methods in Natural Language Processing}, 2667--2678.

\bibitem[{Nguyen and Daum{\'e}~III(2019)}]{nguyen2019hanna}
Nguyen, K.; and Daum{\'e}~III, H. 2019.
\newblock Help, Anna! Visual Navigation with Natural Multimodal Assistance via Retrospective Curiosity-Encouraging Imitation Learning.
\newblock In \emph{Proceedings of the Conference on Empirical Methods in Natural Language Processing (EMNLP)}.

\bibitem[{Pirsiavash and Ramanan(2012)}]{pirsiavash2012detecting}
Pirsiavash, H.; and Ramanan, D. 2012.
\newblock Detecting activities of daily living in first-person camera views.
\newblock In \emph{2012 IEEE conference on computer vision and pattern recognition}, 2847--2854. IEEE.

\bibitem[{Qi et~al.(2020)Qi, Wu, Anderson, Wang, Wang, Shen, and van~den Hengel}]{qi2019rerere}
Qi, Y.; Wu, Q.; Anderson, P.; Wang, X.; Wang, W.~Y.; Shen, C.; and van~den Hengel, A. 2020.
\newblock REVERIE: Remote Embodied Visual Referring Expression in Real Indoor Environments.
\newblock In \emph{Proceedings of the IEEE Conference on Computer Vision and Pattern Recognition (CVPR)}.

\bibitem[{Qiao et~al.(2022)Qiao, Qi, Hong, Yu, Wang, and Wu}]{qiao2022hop}
Qiao, Y.; Qi, Y.; Hong, Y.; Yu, Z.; Wang, P.; and Wu, Q. 2022.
\newblock HOP: History-and-Order Aware Pre-training for Vision-and-Language Navigation.
\newblock In \emph{Proceedings of the IEEE/CVF Conference on Computer Vision and Pattern Recognition}, 15418--15427.

\bibitem[{Radford et~al.(2019)Radford, Wu, Child, Luan, Amodei, Sutskever et~al.}]{radford2019language}
Radford, A.; Wu, J.; Child, R.; Luan, D.; Amodei, D.; Sutskever, I.; et~al. 2019.
\newblock Language models are unsupervised multitask learners.
\newblock \emph{OpenAI blog}, 1(8): 9.

\bibitem[{Rohrbach et~al.(2015)Rohrbach, Rohrbach, Tandon, and Schiele}]{rohrbach2015dataset}
Rohrbach, A.; Rohrbach, M.; Tandon, N.; and Schiele, B. 2015.
\newblock A dataset for movie description.
\newblock In \emph{Proceedings of the IEEE conference on computer vision and pattern recognition}, 3202--3212.

\bibitem[{Schumann and Riezler(2022)}]{schumann2022analyzing}
Schumann, R.; and Riezler, S. 2022.
\newblock Analyzing Generalization of Vision and Language Navigation to Unseen Outdoor Areas.
\newblock \emph{arXiv preprint arXiv:2203.13838}.

\bibitem[{Sharma et~al.(2018)Sharma, Ding, Goodman, and Soricut}]{sharma2018conceptual}
Sharma, P.; Ding, N.; Goodman, S.; and Soricut, R. 2018.
\newblock Conceptual captions: A cleaned, hypernymed, image alt-text dataset for automatic image captioning.
\newblock In \emph{Proceedings of the 56th Annual Meeting of the Association for Computational Linguistics (Volume 1: Long Papers)}, 2556--2565.

\bibitem[{Shridhar et~al.(2020)Shridhar, Thomason, Gordon, Bisk, Han, Mottaghi, Zettlemoyer, and Fox}]{ALFRED20}
Shridhar, M.; Thomason, J.; Gordon, D.; Bisk, Y.; Han, W.; Mottaghi, R.; Zettlemoyer, L.; and Fox, D. 2020.
\newblock {ALFRED: A Benchmark for Interpreting Grounded Instructions for Everyday Tasks}.
\newblock In \emph{The IEEE Conference on Computer Vision and Pattern Recognition (CVPR)}.

\bibitem[{Tan, Yu, and Bansal(2019)}]{tan2019learning}
Tan, H.; Yu, L.; and Bansal, M. 2019.
\newblock Learning to navigate unseen environments: Back translation with environmental dropout.
\newblock \emph{arXiv preprint arXiv:1904.04195}.

\bibitem[{Thomason et~al.(2019)Thomason, Murray, Cakmak, and Zettlemoyer}]{thomason:corl19}
Thomason, J.; Murray, M.; Cakmak, M.; and Zettlemoyer, L. 2019.
\newblock Vision-and-Dialog Navigation.
\newblock In \emph{Conference on Robot Learning (CoRL)}.

\bibitem[{Wang et~al.(2019)Wang, Huang, Celikyilmaz, Gao, Shen, Wang, Wang, and Zhang}]{wang2019reinforced}
Wang, X.; Huang, Q.; Celikyilmaz, A.; Gao, J.; Shen, D.; Wang, Y.-F.; Wang, W.~Y.; and Zhang, L. 2019.
\newblock Reinforced cross-modal matching and self-supervised imitation learning for vision-language navigation.
\newblock In \emph{Proceedings of the IEEE/CVF Conference on Computer Vision and Pattern Recognition}, 6629--6638.

\bibitem[{Wu et~al.(2019)Wu, Kirillov, Massa, Lo, and Girshick}]{wu2019detectron2}
Wu, Y.; Kirillov, A.; Massa, F.; Lo, W.-Y.; and Girshick, R. 2019.
\newblock Detectron2.
\newblock \url{https://github.com/facebookresearch/detectron2}.

\bibitem[{Xiang, Wang, and Wang(2020)}]{xiang2020learning}
Xiang, J.; Wang, X.~E.; and Wang, W.~Y. 2020.
\newblock Learning to stop: A simple yet effective approach to urban vision-language navigation.
\newblock \emph{arXiv preprint arXiv:2009.13112}.

\bibitem[{Xu et~al.(2017)Xu, Zhao, Xiao, Wu, Zhang, He, and Zhuang}]{xu2017video}
Xu, D.; Zhao, Z.; Xiao, J.; Wu, F.; Zhang, H.; He, X.; and Zhuang, Y. 2017.
\newblock Video question answering via gradually refined attention over appearance and motion.
\newblock In \emph{Proceedings of the 25th ACM international conference on Multimedia}, 1645--1653.

\bibitem[{Xu et~al.(2016)Xu, Mei, Yao, and Rui}]{xu2016msr}
Xu, J.; Mei, T.; Yao, T.; and Rui, Y. 2016.
\newblock Msr-vtt: A large video description dataset for bridging video and language.
\newblock In \emph{Proceedings of the IEEE conference on computer vision and pattern recognition}, 5288--5296.

\bibitem[{Zhang, Peng, and Zhou(2022)}]{zhang2learning}
Zhang, Q.; Peng, Z.; and Zhou, B. 2022.
\newblock Learning to Drive by Watching YouTube Videos: Action-Conditioned Contrastive Policy Pretraining.
\newblock \emph{ECCV}, 2(4): 5.

\bibitem[{Zhang and Kordjamshidi(2022)}]{zhang2022explicit}
Zhang, Y.; and Kordjamshidi, P. 2022.
\newblock Explicit Object Relation Alignment for Vision and Language Navigation.
\newblock In \emph{Proceedings of the 60th Annual Meeting of the Association for Computational Linguistics: Student Research Workshop}, 322--331.

\bibitem[{Zhao et~al.(2021)Zhao, Anderson, Jain, Wang, Ku, Baldridge, and Ie}]{zhao2021evaluation}
Zhao, M.; Anderson, P.; Jain, V.; Wang, S.; Ku, A.; Baldridge, J.; and Ie, E. 2021.
\newblock On the evaluation of vision-and-language navigation instructions.
\newblock \emph{arXiv preprint arXiv:2101.10504}.

\bibitem[{Zhou, Liu, and Mu(2021)}]{zhou2021rethinking}
Zhou, X.; Liu, W.; and Mu, Y. 2021.
\newblock Rethinking the Spatial Route Prior in Vision-and-Language Navigation.
\newblock \emph{arXiv preprint arXiv:2110.05728}.

\bibitem[{Zhu et~al.(2021)Zhu, Wang, Fu, Yan, Narayana, Sone, Basu, and Wang}]{zhu-etal-2021-multimodal}
Zhu, W.; Wang, X.; Fu, T.-J.; Yan, A.; Narayana, P.; Sone, K.; Basu, S.; and Wang, W.~Y. 2021.
\newblock Multimodal Text Style Transfer for Outdoor Vision-and-Language Navigation.
\newblock In \emph{Proceedings of the 16th Conference of the European Chapter of the Association for Computational Linguistics: Main Volume}, 1207--1221. Online: Association for Computational Linguistics.

\bibitem[{Zhu et~al.(2015)Zhu, Kiros, Zemel, Salakhutdinov, Urtasun, Torralba, and Fidler}]{Zhu_2015_ICCV}
Zhu, Y.; Kiros, R.; Zemel, R.; Salakhutdinov, R.; Urtasun, R.; Torralba, A.; and Fidler, S. 2015.
\newblock Aligning Books and Movies: Towards Story-Like Visual Explanations by Watching Movies and Reading Books.
\newblock In \emph{The IEEE International Conference on Computer Vision (ICCV)}.

\end{thebibliography}
